\documentclass[10pt,twocolumn,letterpaper]{article}

\usepackage{cvpr}              

\definecolor{cvprblue}{rgb}{0.21,0.49,0.74}
\usepackage[pagebackref,breaklinks,colorlinks,allcolors=cvprblue]{hyperref}
\usepackage{lineno}

\def\paperID{14696} 
\def\confName{CVPR}
\def\confYear{2025}

\title{DIV-FF: Dynamic Image-Video Feature Fields\\ For Environment Understanding in Egocentric Videos}

\author{Lorenzo Mur-Labadia\\
{\tt\small lmur@unizar.es}
\and
Josechu Guerrero\\
{\tt\small jguerrer@unizar.es}
\and
Ruben Martinez-Cantin\\
{\tt\small rmcantin@unizar.es}
}

\begin{document}

\twocolumn[{%
\renewcommand\twocolumn[1][]{#1}%
\maketitle
\vspace{-2em}
\begin{center}
    \captionsetup{type=figure}
    \includegraphics[width=0.9\linewidth]{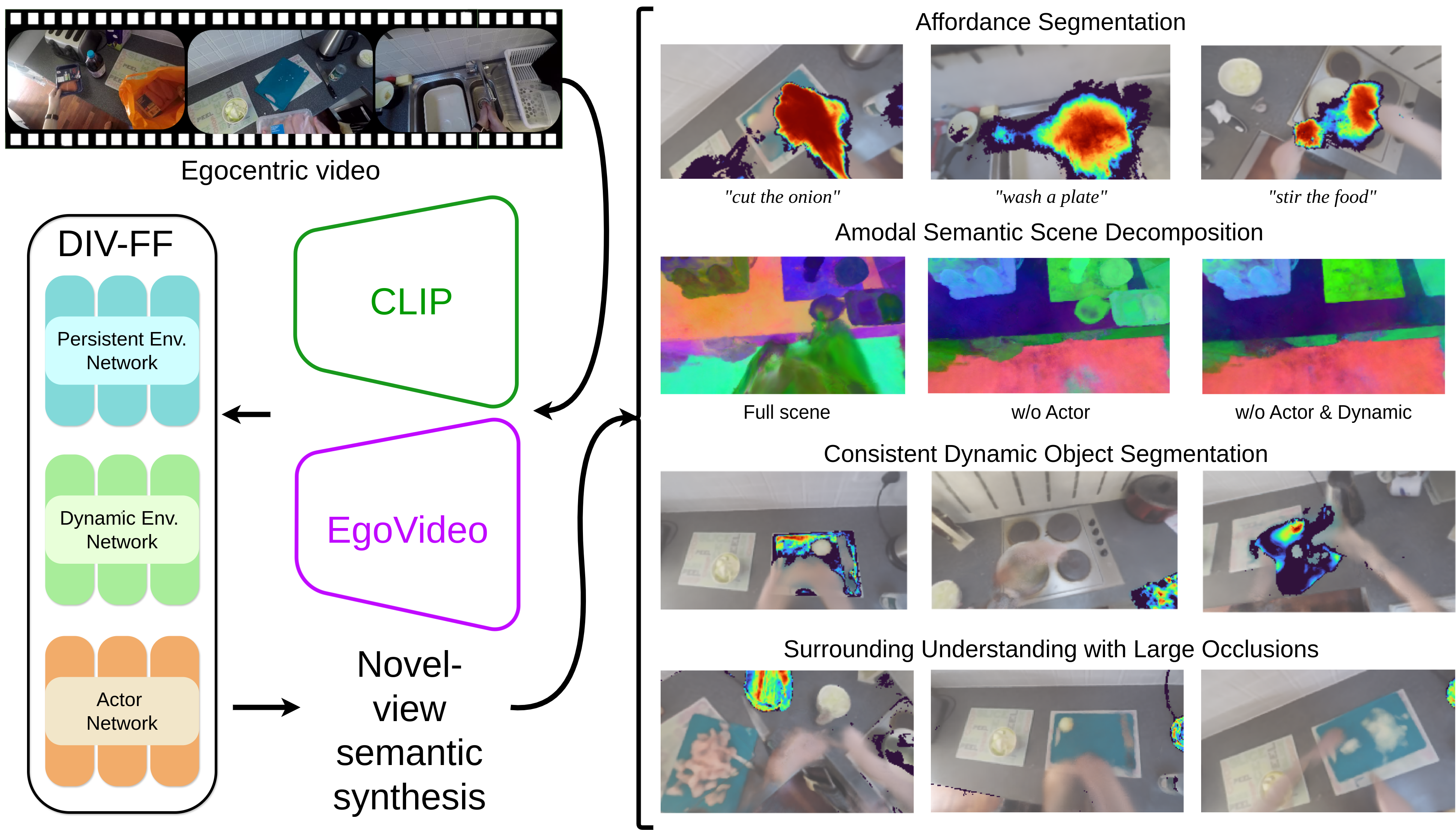}
    \captionof{figure}{DIV-FF distills image and video language features in a triple stream feature field tailored to egocentric videos with numerous interactions and camera wearer movements. Our approach achieves a deep understanding of the environment, supporting precise affordance segmentation, semantic scene decomposition and consistent segmentation of dynamic objects. With its implicit 3D representation, DIV-FF comprehends not just novel views but also surrounding areas.}
    \label{fig:splash_fig}
\end{center}

}]

\definecolor{verdeOscuro}{rgb}{0.0, 0.5, 0.0} 

\begin{abstract}

Environment understanding in egocentric videos is an important step for applications like robotics, augmented reality and assistive technologies. These videos are characterized by dynamic interactions and a strong dependence on the wearer’s engagement with the environment. Traditional approaches often focus on isolated clips or fail to integrate rich semantic and geometric information, limiting scene comprehension. We introduce Dynamic Image-Video Feature Fields (DIV-FF), a framework that decomposes the egocentric scene into persistent, dynamic, and actor-based components while integrating both image and video-language features. Our model enables detailed segmentation, captures affordances, understands the surroundings and maintains consistent understanding over time. DIV-FF outperforms state-of-the-art methods, particularly in dynamically evolving scenarios, demonstrating its potential to advance long-term, spatio-temporal scene understanding.

\end{abstract}
    
\section{Introduction}
\label{sec:intro}




Egocentric videos offer a unique way to understand human activities from a first-person perspective, benefiting applications like mobile robotics, augmented reality and assistive devices. 
In these videos, an actor continuously moves to interact sporadically with multiple dynamic objects in a static scene, breaking the usual rigid scene assumption.
This tight integration between objects, actions and the dynamic scene introduces both opportunities and challenges for environmental understanding from egocentric videos.


Most existing methods in egocentric environment understanding either consider a short video clip isolated from the physical space \cite{ma2016going, furnari2019would, sudhakaran2019lsta, girdhar2021anticipative, mur2024aff} or they provide a strong spatial representation but with low semantic understanding \cite{park2016egocentric, rhinehart2017first, rhinehart2016learning, liu2022egocentric}. 
However, when humans interact repeatedly in a fixed environment, we develop a physical and semantic model that integrates the spatial distributions of the elements around us, both \emph{persistent} and \emph{dynamic}.
The semantics capture detailed information about \emph{objects} and their \emph{attributes} through natural language descriptions. 
Additionally, we encode the available action (i.e.: \emph{affordance}) locations in the environment, linking physical zones of interaction to the likely activities they support.
Besides, we dynamically update this persistent semantic model as we interact, recording the location and state of dynamic objects at every moment.
In that sense, some approaches propose intermediate representations between a pure semantic understanding of the video without explicit representation and a pure geometrical representation. They adopt semantic topological maps \cite{nagarajan2020ego}, local environment state representations \cite{nagarajan2024egoenv} or explicit representations \cite{plizzari2024spatial, mur2023multi} for improving the environment understanding of the scene. In this work, we build an implicit (neural network) model that is able to jointly capture the geometry, appearance and semantic understanding encoded in the video, and enable predictions in novel-view points using Neural Radiance Fields (NeRFs).


NeRFs provide a compact implicit representation of the geometry and visual appearance of a scene \cite{mildenhall2021nerf}. The implicit representation of NeRFs can also be used for semantic encoding, supporting multiple applications like robot manipulation \cite{kerr2024robot, lou2024robo}, navigation \cite{werby2024hierarchical}, or scene editing \cite{kobayashi2022decomposing,ren2024nerf}. 
For example, Neural Feature Fusion Fields (N3F) \cite{tschernezki2022neural} extends the NeRFs predictive capabilities in a teacher-student fashion, where a teacher model that predicts semantic features in image space is used to train a NeRF-like student to predict semantic features in 3D space. 
These semantic capabilities are further extended in Language Embedded Radiance Fields (LERF) \cite{kerr2023lerf},
enabling natural language query in 3D locations by volume rendering CLIP embeddings. However, LERF assumes a rigid scene which limits its applicability to egocentric videos where the actor is interacting with the environment. Furthermore, semantic distillation is based on single-image semantic features (e.g., CLIP features) which do not capture the dynamic nature of actions or changing elements.


In this work, we propose DIV-FF (Dynamic Image-Video Feature Fields
), the first language embedded feature field capable of decomposing both the geometry and the semantics of the scene for the actor, and also for the persistent and dynamic elements via three different streams. 
While previous works focus on image-language embeddings, we also introduce video-language embeddings (based on EgoVideo \cite{pei2024egovideo}) to understand fine-grained action descriptions. This encodes the environment affordances, possible actions available in the environment for the actor, linking specific activities to physical zones where interactions are likely to occur.
A parallel feature field, based on image-language features from CLIP, captures detailed information about objects and their attributes, categorizing them through natural language descriptions rather than fixed semantic tags, even from novel viewpoints. Its implicit representation, similar to NeRFs, ensures that even areas not visible from the egocentric camera remain strongly connected in the environment model.
Although this environment model provides a persistent long-term representation, it is dynamically updated as the user interacts, enabling a precise record of the location and state of dynamic objects at every moment. 
Our main \textbf{contributions} are as follows:
\begin{itemize}
    \item We distill video-language embeddings (from EgoVideo) to understand temporally dependent semantics, such as affordances (available actions), which single-image models like CLIP cannot capture. 
    \item We propose an approach to adapt 
    language embedded feature fields to dynamic egocentric videos by dividing the radiance and feature fields depending on whether they are from the actor, dynamic, or persistent elements.
    \item We present a robust image-language feature field enhanced by leveraging SAM masks, which also includes the temporal dependency and achieves a consistent segmentation of the dynamic objects over time. 
    \item Our results demonstrate significant improvements in dynamic object (+40.5$\%$) and affordance segmentation (+69.7 $\%$) by using text query relevancy maps. Furthermore, our model effectively connects the egocentric view with the semantics of the surroundings and decomposes the scene into different levels.
    
\end{itemize}

\section{Related works}

\noindent\textbf{Egocentric environment understanding using geometric representations.}
Some works that consider the physical layout build semantic explicit representations from videos of indoor scenes using visual SLAM systems. 
Rhinehart et al. \cite{rhinehart2016learning} learn 2D maps with the functionality of different actions.
Semantic MapNet \cite{henriques2018mapnet} propose a birds-eye-view spatial memory for mapping, which is updated with recurrent neural networks to remember places visited in the past.
Cartillier et al. \cite{cartillier2021semantic} encode the egocentric frame, project its features, and then decode the semantic labels in a 2D map.
Liu et al. \cite{liu2022egocentric} recognize and localize activities in an existing 3D voxel map from an egocentric video. 
The limitations in extracting the camera pose from egocentric video \cite{patra2019ego} due to the quick camera movements and motion blur have hampered the unification of 3D geometry and video understanding.
Recently, the arrival of egocentric 3D datasets with camera poses \cite{mur2023multi, tschernezki2024epic, grauman2024ego, xia2018gibson} and the improvement of  3D sensors like project ARIA \cite{engel2023project} has unlocked the arrival of novel works.
Plizzari et al. \cite{plizzari2024spatial} track active objects through their appearance and spatial consistency in the 3D scene, even when they are out of view.
Mur-Labadia et al. \cite{mur2023multi} extract 2D affordance segmentation maps to build later a point cloud of the environment encoding those labels. 
Tschernezki et al. \cite{bhalgat20243d} proposed a 3D-aware instance object tracking by keeping a long-term consistency.
EgoLoc \cite{mai2023egoloc} extend episodic memory to 3D by estimating the relative 3D object pose to the user.




\noindent\textbf{Egocentric environment understanding without geometric representations.} 
Most egocentric video understanding works just consider a short time window of the video. Although these works obtain a remarkable semantic understanding in multiple tasks like action recognition \cite{ma2016going, sudhakaran2019lsta}, object segmentation \cite{shan2020understanding, tschernezki2024epic}, action forecasting \cite{furnari2019would, mur2024aff} or capturing activity threads \cite{price2022unweavenet}, these approaches ignore the underlying physical space of the scene. Some approaches \cite{nagarajan2020ego, nagarajan2024egoenv, ramakrishnan2022environment} extract environment-aware features via alternative representations that avoid the geometric reconstruction problems from SLAM in egocentric videos \cite{patra2019ego}.
Ego-Topo \cite{nagarajan2020ego} builds a topological map, where the nodes represent environment zones with a coherent set of interactions linked by their spatial proximity. 
EgoEnv \cite{nagarajan2024egoenv} encodes the relative directions of the objects to the camera wearer in a local state vector, learning an environment-aware video representation.
Ramakrishnan et al. \cite{ramakrishnan2022environment} capture the inherent statistics of indoor environments to learn an environment predictive coding, which applies later for navigation.

\noindent\textbf{Dynamic Radiance Fields.}
Neural Radiance Fields (NeRFs) \cite{ren2024nerf} allow capturing and rendering complex 3D scenes from a set of multi-view posed images. Using an implicit function and via differentiable volume rendering, NeRFs map spatial coordinates and viewing directions to colors and densities. 
Early methods for rendering dynamic scenes \cite{li2021neural, gao2021dynamic} use pre-trained motion segmentation methods to mask moving objects, guiding separate NeRFs networks to disentangle motion-based components.
Liang et. al \cite{liang2023semantic} leverage DINO features to identify salient foreground regions along spacetime, while Wu et al. \cite{wu2022d} decouple moving objects from the static background in a self-supervised manner with two neural radiance fields.
NeuralDiff \cite{tschernezki2022neural} separates the static background, dynamic objects and the actor's body via inductive biases, obtaining a different implicit representation for each part of the scene.
Recently, Zhang et al. \cite{zhang2024egogaussian} optimize 3D Gaussians to reconstruct the scene and track the 3D object motions from an egocentric video, but requiring pre-extracted hand-object interaction masks.

\begin{figure*}[t!]
    \includegraphics[width=\textwidth]{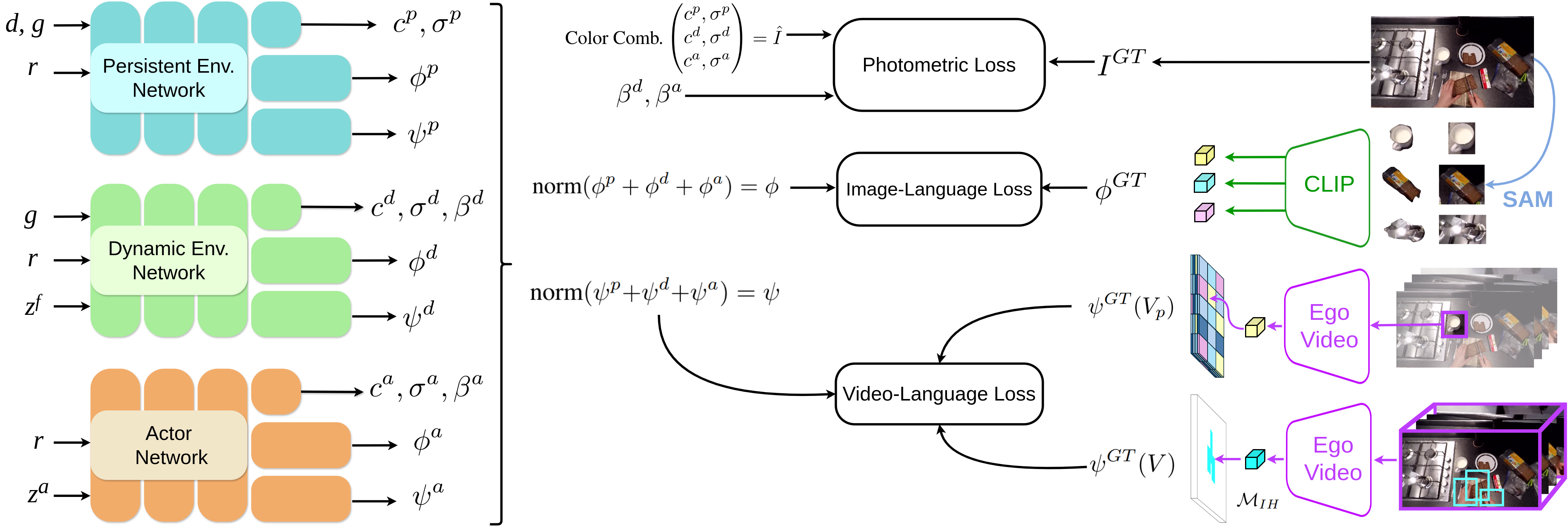}
    \caption{\textbf{Overview of DIV-FF.} Our three-stream architecture field predicts the color $c$, the density $\sigma$, the material aleatoric uncertainty $\beta$, the image-language features $\phi$ and the video-language features $\psi$ along a ray $r$ with direction $d$ given the camera viewpoint $g$ and a frame specific code $z$. 
We first extract SAM masks and bounding boxes from the image, that we leverage to obtain a unique CLIP descriptor $\phi_{GT}$ in all the pixels within the respective mask.
We supervise the video-language feature field with local patch features $\psi^{GT}(V_p)$ and a global video embedding $\psi^{GT}(V)$ assigned only to pixels in the interaction hotspot $\mathcal{M}_{IH}$, computed with a pre-trained hand-object detector.}
    \label{fig:architecture}
\end{figure*}

\noindent\textbf{Feature Distillation in NeRFs.} Several works extend radiance fields to integrate 2D semantic labels into the 3D space during the optimization \cite{zhi2021place, vora2021nesf, siddiqui2023panoptic, fu2022panoptic}. In contrast, the objective of 3D feature distillation methods \cite{tschernezki2022neural, goel2023interactive, kobayashi2022decomposing, yue2025improving} is to transfer 2D image features from a teacher model (i.e, a self-supervised model like DINO \cite{caron2021emerging}) into a 3D student neural renderer.
Expanding on this, 3D language feature fields distill image-text CLIP features \cite{radford2021learning}, enabling querying the 3D student with open-vocabulary text descriptions to obtain relevancy maps.
LERF \cite{kerr2023lerf} fuses multi-scale patch-level CLIP features conditioned on the scale.
N2F2 \cite{bhalgat2024n2f2} addresses the need for evaluating the rendering at the different scales by learning a unified feature field, where the different semantic granularities are encoded in a high-dimensional feature space.
LangSplat \cite{qin2024langsplat} adopts 3D Gaussians \cite{kerbl20233d} and combines CLIP features with multi-scale SAM masks, improving the segmentation quality.
EgoLifter \cite{gu2025egolifter} augments 3D Gaussian Splatting with instance features from egocentric videos, but it only reconstructs the static part of the scene by filtering out the actor and the dynamic objects.

\section{Dynamic Image and Video Feature Fields (DIV-FF) from Egocentric Videos}

Our approach is to build a language embedding feature field that decomposes the 3D representation in three components (persistent environment, dynamic environment and actor) for accounting the inherent dynamics present in egocentric videos. 
In addition, we incorporate a second modality stream of embeddings based on video-language models which can capture the action semantics only present in the video modality. 
Besides, we introduce a time-dependent module on the dynamic and actor stream, capturing the temporal evolution of the feature fields. 

\subsection{Dynamic Neural Radiance Fields}

The geometry model \cite{tschernezki2021neuraldiff} captures the dynamic scene by integrating three different radiance fields, illustrated in Figure \ref{fig:architecture}.
The \textit{persistent environment network} predicts the color $c^p_k$ and density $\sigma^p_k$ at each point along a ray $r_k$, given a viewing position $g_t$ and unit-norm viewing direction $d_t$. Formally, it is defined as $(c^p_k, \sigma^p_k ) = \text{MLP}^p(g_t r_k, d_t)$.
To model the dynamic objects in the scene, a second \textit{dynamic environment network} $(c^d_k, \sigma^d_k , \beta^d_k) = \text{MLP}^d(g_t r_k, z^d_t)$ estimates the density $\sigma^d_k$ and the color as a Gaussian distribution $\mathcal{N}(c^d_k, \beta^d_k$), where $\beta^d_k$ represents the heteroscedastic aleatoric uncertainty associated with the color. It also includes as input a frame-specific code $z^d_t$ that accounts for temporal variations of the dynamic objects, which exhibit sporadic motion relative to the global reference frame.
The \textit{actor network} is similar to the dynamic environment network, but since the actor moves continuously linked to the camera, it removes the projection of the ray to the world coordinate system $(c^a_k, \sigma^a_k , \beta^a_k) = \text{MLP}^a(r_k, z^a_t)$.
Here, $z^a_t$ is a frame-specific parameter designed to capture the continuous motion of the actor.
The predicted material uncertainty terms $\beta^d_k, \beta^a_k$ indicate the confidence levels associated with each ray $r_k$ for representing the dynamic objects and the actor, respectively. By employing improved color mixing techniques and inductive biases during training, the model accurately reconstructs scene dynamic geometry as a composite of the three radiance fields. For more details on the geometric model, please refer to \cite{tschernezki2021neuraldiff}.

\subsection{Image-Language Feature Field}
\label{sec:ilff}

We extend the three-stream geometry model to distill image-language semantic features from CLIP \cite{radford2021learning}.
Since the CLIP image encoder is a global image descriptor, it lacks pixel-aligned embeddings. To address this, LERF \cite{kerr2023lerf} extracts multi-scale patch-level features, which often fail to encompass the target object or add excessive contextual information. It results in blurred object boundaries and noise, requiring DINO \cite{caron2021emerging} for regularization.

As shown in Figure \ref{fig:architecture}, our model incorporates pixel-aligned CLIP features by leveraging accurate object masks generated by Segment Anything Model (SAM) \cite{kirillov2023segment} inspired by recent works \cite{qin2024langsplat, bhalgat2024n2f2}.
Specifically, we extract CLIP features per each segmented mask region $\phi^{GT}_\mathcal{M}$ and its respective bounding box $\phi^{GT}_{\mathcal{B}}$. We assign the same weighted descriptor $\phi^{GT}= 0.75 \cdot \phi^{GT}_\mathcal{M} + 0.25 \cdot \phi^{GT}_{\mathcal{B}}$ to all the pixels within the mask. This balanced approach achieves pixel-level alignment while preserving semantic context.
Furthermore, the use of precise semantic masks with sharp object boundaries eliminates the need for DINO regularization used in previous works \cite{kerr2023lerf}.



\begin{table*}[t]
\centering
\resizebox{\textwidth}{!}{%
\begin{tabular}{l|cccccccccc|c}
Method & S01 & S02 & S03 & S04 & S05 & S06 & S07 & S08 & S09 & S10 & Average mIoU \\ \hline
LERF & 22.1 & 10.1 & 11.7 & 9.7 & 13.2 & 18.6 & 12.5 & 6.2 & 19.0 & 5.5 & 12.8 \\ 
NeuralDiff + OWL-ViT & 9.4 & 10.2 & 13.2 & 15.4 & 9.4 & 13.3 & 14.5 & 23.2 & 23.7 & 28.9 & 16.1 \\
NeuralDiff + OWL-ViT + SAM & 8.7 & 12.6 & 23.2 & 23.9 & 13.8 & 15.9 & 17.8 & \textbf{28.0} & \textbf{32.9} & \textbf{41.1} & \underline{21.7} \\ \hline
DIV-FF (CLIP in patches) & 26.9 & 21.7 & 18.3 & 16.8 & 18.1 & 24.9& 17.3 & 12.3 & 17.9 & 23.6 & 19.8\\
DIV-FF (CLIP in SAM masks)  & 30.7 & 19.3 & 29.6 & 24.9 & \textbf{31.3} & 26.1 & 28.8 & 14.8 & 23.8 & 35.1 & 26.2\\ \hline
DIV-FF (full model, video infer.)& 16.1 & 15.4 & 9.3  & 9.5  & 21.8 & 20.7 & 10.7 & 18.5 & 17.9 & 20.6 & 16.6 \\ \hline
DIV-FF (full model, image infer.) &  \textbf{40.3}&  \textbf{30.4}&  \textbf{37.4}&  \textbf{29.8}&  29.5& \textbf{32.6}& \textbf{30.6}& 15.1 &  25.1&  33.6& \textbf{30.5} {\color{verdeOscuro}{(+40.5$\%$)}} \\ \hline
\end{tabular}%
}
\caption{\textbf{Dynamic Object Segmentation by CLIP image-language feature field.} 
Compared with LERF, DIV-FF considers a dynamic scene in the geometric reconstruction. Our full model assigns the same descriptor to all the pixels within a SAM mask. This descriptor is a weighted average between the CLIP of the mask and the bounding box. We compute relative improvement against the \underline{best baseline model}.}
\label{tab:object_segm}
\end{table*}

\subsection{Video-Language Feature Field}
\label{sec:vlff}

While the CLIP image-language features contain fine-grained and accurate details of the objects, they ignore interaction semantics present in egocentric videos as they require temporal information. Therefore, we incorporate in parallel a video-language feature field to capture dynamic semantics, such as \emph{affordances} and potential interactions. 
We leverage Video-Language Pre-trained (VLP) models \cite{pramanick2023egovlpv2, pei2024egovideo}, which offer richer and action-oriented context by pairing narrative descriptions with video using contrastive learning. We select Ego-Video \cite{pei2024egovideo}, the state-of-the-art in multiple Ego4D \cite{grauman2022ego4d} challenges, for this task. 
Similar to CLIP, the video encoder of Ego-Video outputs a single descriptor from a video patch, not pixel-aligned features. 
In this case we cannot use object masks as in Section \ref{sec:ilff}, because our goal is to identify \emph{interaction hotspot} regions, including both the hands and the object parts (e.g. ``knife edge'', ``spatula handle''), not just the entire object. While SAM’s small masks could localize these areas, their limited size loses essential action context.

Therefore, we distill the video-language feature field with patch and global-level embeddings.
We first pre-compute video descriptors $\psi^{GT}(V_p)$ from medium-sized video patches $V_p$, balancing fine-grained details with action context. Second, we derive a global descriptor $\psi^{GT}(V)$ for the entire video, assigned solely to the pixels within the interaction hotspot area $\mathcal{M}_{IH}$ \cite{nagarajan2019grounded} and combine them in a single loss function:
\begin{equation}
    \mathcal{L}_{V} =   \left\| \hat{\psi} - \psi^{GT}(V_p)\right\|^2 + \mathcal{M}_{IH} \left\| \hat{\psi} - \psi^{GT}(V)\right\|^2
\end{equation}
This improves the feature field's capability to capture relevant interaction regions.
We obtain the interaction hotspot mask $\mathcal{M}_{IH}$ as the union of the hands and active objects bounding box, pre-extracted with an existing hands-object detector \cite{shan2020understanding}. 
Additionally, the training of this feature field is regularized with pixel-aligned DINO \cite{caron2021emerging} features thanks to its object decomposition properties \cite{kerr2023lerf}.

\section{Experimental Settings}

\begin{figure*}[h!]
    \centering
    \includegraphics[width=0.99\textwidth]{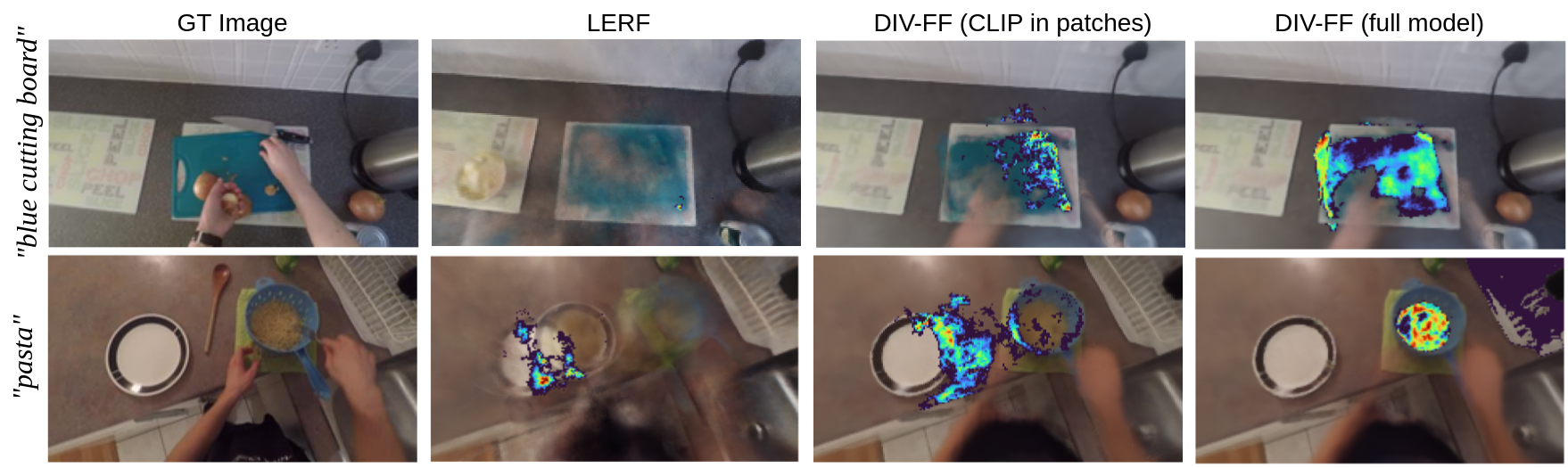}
    \caption{\textbf{Ablations on the image-language feature field.} Treating the egocentric video as a dynamic scene enhances geometric reconstruction, while utilizing SAM masks further improves object segmentation accuracy.}
    \label{fig:comparative}
\end{figure*}

\begin{figure*}[h!]
    \centering
    \includegraphics[width=0.99\textwidth]{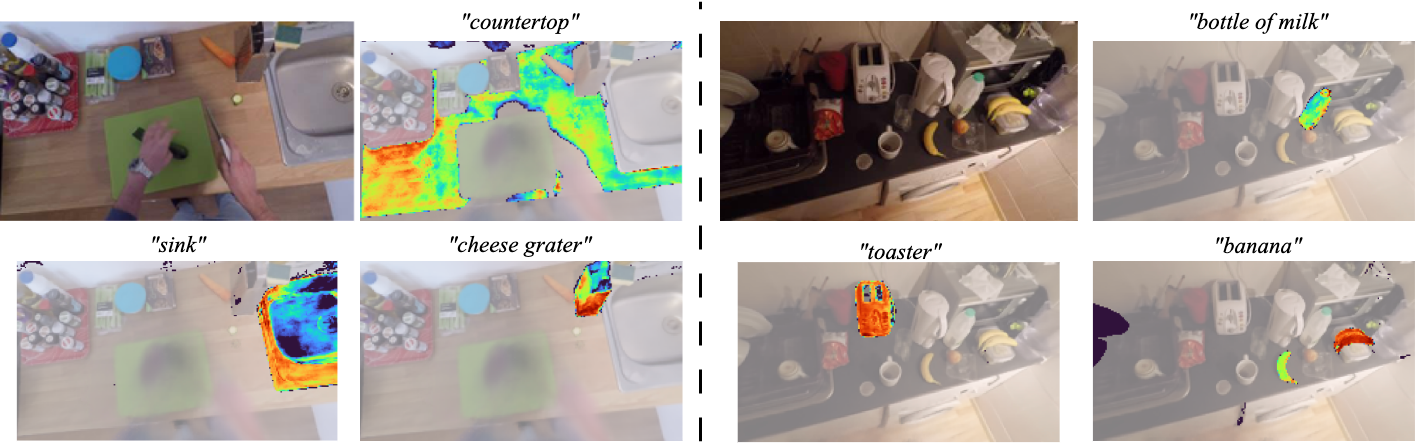}
    \caption{\textbf{DIV-FF Image-Language relevancy maps in novel-views.} We can see the performance of various text queries for dynamic object segmentation. We can see how the object contours are well defined as we used masks during training.}
    \label{fig:object_Segm}
\end{figure*}


\noindent\textbf{Implementation details.} We extend the three stream architecture of NeuralDiff \cite{tschernezki2021neuraldiff} by incorporating 4-layer, 256-width MLPs for the image $\phi$ and video language $\psi$ feature fields, respectively. Both the coarse and fine models use 64 samples, while we select the best 32 samples for the feature distillation. The representations are summed and normalized post-rendering. 
We use an Adam optimizer with a learning rate of 5 $\times 10^{-4}$ and a cosine annealing scheduler. We train the geometry for 10 epochs with a batch size of 1024, then distill semantic features in two phases: training only the semantic heads for 5,000 iterations, followed by the full model for 3 epochs on an NVIDIA 4090.

\noindent\textbf{Feature extractors.} To extract the CLIP image embeddings, we utilize the OpenCLIP ViT-B/16 model \cite{cherti2023reproducible} trained on the LAION-2D dataset following \cite{kerr2023lerf} for fair comparison.
We prompt SAM \cite{kirillov2023segment} with a 32 $\times$ 32 grid, filtering redundant masks by 0.7 IoU, 0.85 stability score, and 0.7 overlap rate. 
We reduce the dimensionality of CLIP descriptors by training a scene-wise language auto-encoder \cite{qin2024langsplat} to reduce the memory cost. 
We sample 4 video frames at 60 fps with a temporal stride of 15. Local-patch video features are extracted using a patch size of 33\% the image size and a stride factor of 0.5. For masking the global video descriptor, we employ a hand-object detector \cite{shan2020understanding}, specifically finetunned for egocentric sequences.

\noindent\textbf{Dataset.} We conduct our experiments on the EPIC-Diff subset \cite{tschernezki2021neuraldiff} of the EPIC-Kitchens \cite{damen2018scaling} dataset.
On average, each sequence comprises of 900 calibrated frames, spanning 14 minutes of egocentric video, featuring multiple viewpoints and a large number of manipulated objects. 
Our evaluation encompasses both our method and the baselines on the test set, which includes frames not utilized during model training. This set facilitates assessments of new-view synthesis and segmentation capabilities.

\noindent\textbf{Baselines.} We compare against the following baselines:
\begin{itemize}
    \item \textbf{LERF} \cite{kerr2023lerf} assumes a static scene and uses a single stream for geometry and semantics. It employs multi-resolution hash-grids to optimize both the radiance field and the CLIP language field, which is distilled from multi-scale patch-level CLIP features.
    \item \textbf{OWL} \cite{minderer2024scaling}. We apply this open-vocabulary object detector on the novel-view rendered images produced by Neural-Diff \cite{tschernezki2021neuraldiff}.
    \item \textbf{OWL+SAM} \cite{minderer2024scaling, kirillov2023segment} obtains the object's masks from the bounding box coordinates provided by the OWL baseline.
\end{itemize}

\noindent \textbf{Ablations.} We compare different versions of DIV-FF.
\begin{itemize}
    \item \textbf{DIV-FF (CLIP in patches)} keeps the CLIP patch features from LERF $\phi^{GT}=\phi^{GT}_\mathcal{P}$, but it introduces the dynamic geometry model from \cite{tschernezki2021neuraldiff}.
    \item \textbf{DIV-FF (CLIP in SAM masks)} substitutes CLIP patch features by embeddings from SAM masks  $\phi^{GT} =  \phi^{GT}_\mathcal{M}$.
    \item \textbf{DIV-FF (full model, image inference)} incorporates the bounding box to obtain the CLIP descriptor $\phi^{GT} = 0.75 \cdot \phi^{GT}_\mathcal{M} + 0.25 \cdot \phi^{GT}_\mathcal{B}$, where $\phi^{GT}_\mathcal{B}$.
    \item \textbf{DIV-FF (full model, video inference).} In the full model, we render from the parallel video-language feature field.  
\end{itemize}

\begin{figure*}[ht]
    \centering
    \includegraphics[width=0.99\textwidth]{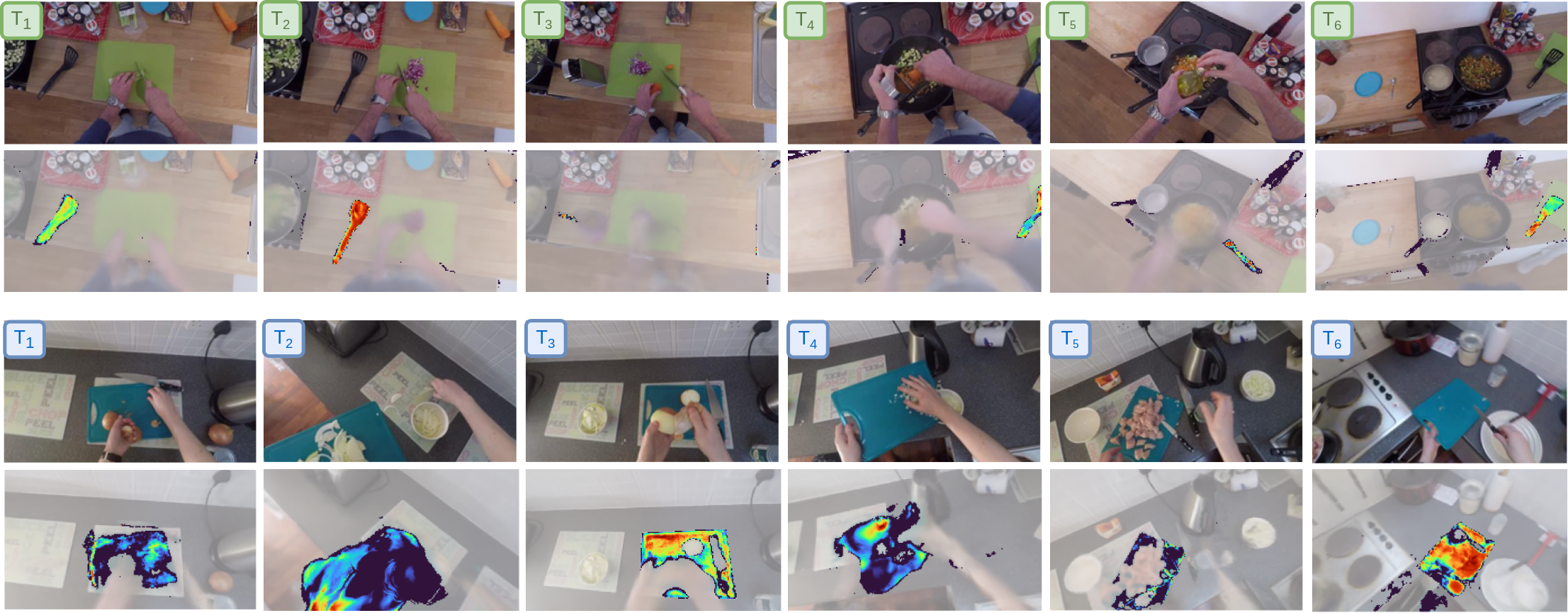}
    \caption{\textbf{Consistent Dynamic Object Segmentation along different time-steps in novel views:} The dynamic and actor streams contain respective frame-specific codes $z^f_t$ and$z^a_t$. This time encoding is also propagated to the semantic feature field, obtaining consistent segmentations despite the continuous movement of the \textit{``spatula''} and \textit{``blue cutting board''}.}
    \label{fig:tracking}
\end{figure*}

\begin{figure}[ht]
    \centering
    \includegraphics[width=0.99\linewidth]{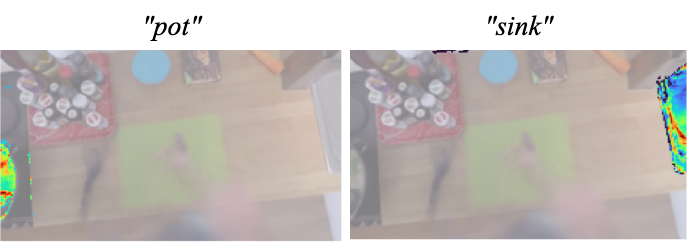}
    \caption{\textbf{Surrounding Understanding.} DIV-FF understands the novel view and the surrounding environment, enabling segmentation of objects at the image's edges with limited observability.}
    \label{fig:env_segm}
\end{figure}


\section{Results}

Once trained, our DIV-FF model predicts the color $c$, density $\sigma$, CLIP $\phi$ and EgoVideo $\psi$ semantic features of a novel-view in an specific time-step and separates the actor and the dynamic elements  from the persistent environment. We evaluate this comprehensive spatio-temporal semantic understanding in different downstream tasks.

\begin{figure*}[t]
    \centering
    \includegraphics[width=0.99\textwidth]{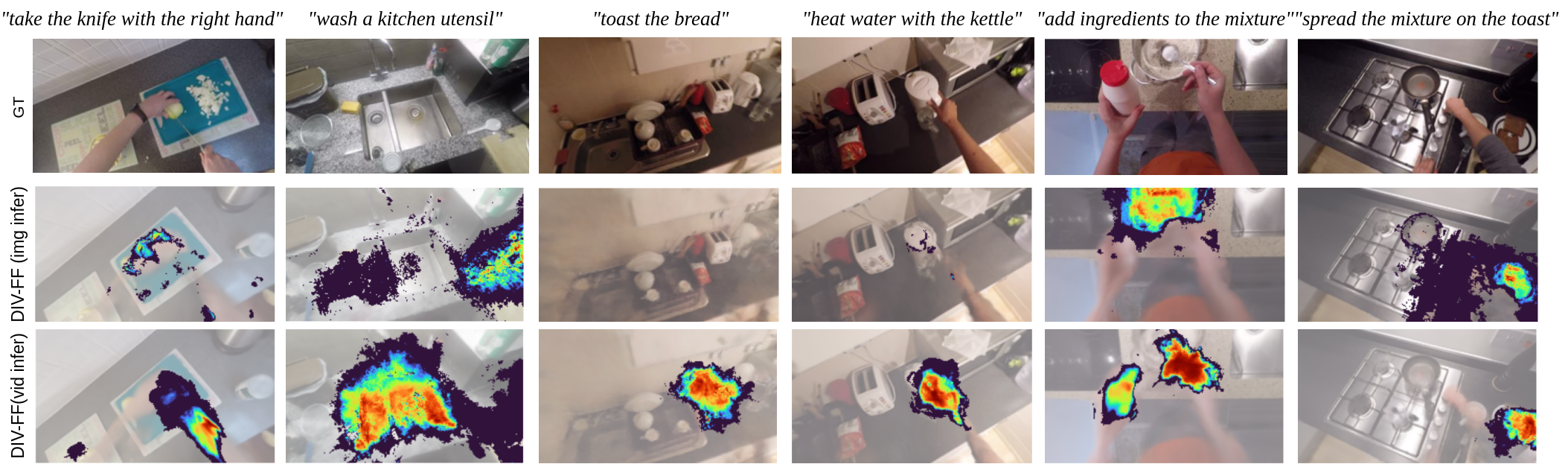}
    \caption{\textbf{Affordance Segmentation qualitative examples.} We compare the relevancy maps produced by the image-language field against those from the video-language field of DIV-FF, based on a detailed action description text query.}
    \label{fig:aff_seg}
\end{figure*}

\subsection{Dynamic Object Segmentation} 

In each scene, we identify a subset of objects that move throughout the video and evaluate in the novel-views the relevancy maps originated by the text queries in the $\phi_{img}$ image-language feature field.
Following the method proposed by LERF \cite{kerr2023lerf}, we compute the relevancy score as: 
$\min_i \frac{\exp (\phi_{img} \cdot \phi_{quer})}{\exp(\phi_{img} \cdot \phi_{quer}) + \exp (\phi_{img} \cdot \phi^i_{can})}$.
This formula evaluates how closely the rendered embedding $\phi_{img}$ matches the query embedding $\phi_{quer}$ compared to a set of predefined canonical phrases $\phi^i_{can}$ (\textit{``object'', ``thing'', ``stuff'', ``texture'', ``hands''}). Segmentation masks are generated for relevance scores above a specified threshold. For the evaluation, we leverage existing annotations from \cite{tschernezki2022neural} and report the mean intersection over union (mIoU). We visualize the text query relevancy maps by normalizing from 50 $\%$ to the maximun relevancy.

Table \ref{tab:object_segm} presents quantitative results on EPIC-Diff scenes.
The full version of DIV-FF achieves the best performance (30.5 mIoU), surpassing the OWL+SAM detector (21.7 mIoU) by +40.5$\%$, illustrating that distilling semantic features outperforms traditional open vocabulary object detection from novel views, since the OWL model fails due to artifacts and the blurry hand effects in the novel view rendered. 
The CLIP patch-level version of DIV-FF (19.8 mIoU) significantly improves upon LERF by explicitly considering the dynamics parts in the semantic and geometric fields with the triple stream architecture of DIV-FF. This leads to sharper reconstructions, particularly for moving objects as shown in Figure \ref{fig:comparative}.
Subsequently, leveraging SAM to extract object-level CLIP features further improves performance (26.2 mIoU), and generates more accurate and consistent semantic renderings compared to CLIP patch-level embeddings. Finally, the introduction of contextual information from the object bounding boxes ultimately yields to the best performance (30.5 mIoU).

Figure \ref{fig:object_Segm} showcases novel-view renderings for various text queries in two scenes, effectively capturing fine-grained details like the \textit{``countertop''} borders. The uniform assignation of the same CLIP descriptor across all object pixels allows DIV-FF to segment objects of any size, such as \textit{``sink''} or \textit{``banana''}. As Figure \ref{fig:tracking} shows, our approach also segments consistently dynamic objects across multiple novel views in different time-steps of the sequence due to the combined impact of object-level CLIP features and the temporal encoding in the frame-specific codes. 
Unlike egocentric methods limited to short time windows, our environment understanding extends beyond the current view to the surrounding regions.
Figure \ref{fig:env_segm} illustrates this capability, showing how our 3D semantic implicit model segments the \textit{``pot''} and \textit{``sink''}, despite being almost occluded in the edge of the image.

\begin{table*}[t]
\centering
\resizebox{\textwidth}{!}{%
\begin{tabular}{l|cccccccccc|c}
Method & S01 & S02 & S03 & S04 & S05 & S06 & S07 & S08 & S09 & S10 & Average mIoU \\ \hline
OWL-ViT & 4.8 & 4.2 & 1.4 & 5.6 & 4.8 & 2.3 & 13.2 & 4.0 & 5.4 & 4.4 & 5.0 \\
OWL-ViT + SAM & 4.6 & 5.4 & 1.9 & 4.6 & 5.8 & 1.1 & 8.6 & 4.5 & 7.7 & 8.3 & 5.3 \\
LERF & 18.2 & 17.4 & 6.8 & 11.5 & 11.9 & 18.4 & 11.7 & 7.5 & 15.2 & 4.2 & \underline{12.2} \\ \hline
DIV-FF (CLIP in patches) & 17.1 & 15.6 & 7.1 & 9.4 & 12.9 & 19.7 & 12.4 & 11.3 & 15.3 & 12.6 & 13.3 \\ \hline
DIV-FF (full m., image infer.) & 17.3 & 13.7 & 6.2 & 13.7 & 19.1 & 8.1 & 18.5 & 7.1 & 11.1 & 3.6 & 11.8 \\ 
DIV-FF (full m., video infer.) & \textbf{20.6} & \textbf{19.9} & \textbf{14.4} & \textbf{22.4} & \textbf{30.1} & \textbf{22.3} & \textbf{20.1} & \textbf{16.8} & \textbf{17.1} & \textbf{23.1} & \textbf{20.7} {\color{verdeOscuro}{(+69.7$\%$)}} \\ \hline
\end{tabular}%
}
\caption{\textbf{Affordance Segmentation.} We compare the segmentation masks of a set of affordable actions in the scene. The full version of DIV-FF is composed by two parallel semantic feature fields, image (CLIP + SAM + boxes descriptors) and video (Ego-Video) respectively.
We compute relative improvement against the \underline{best baseline model}.}
\label{tab:aff_segm}
\end{table*}

\subsection{Affordance Segmentation}
We identify affordable actions in each scene and generate Ego-Video  \cite{pei2024egovideo} text queries $\psi_{quer}$ describing the interaction, which are more complex than simple object labels as they capture nuanced action dynamics.
We compute the relevancy score from the video-text feature field $\psi$ against a different set of canonical phrases $\psi^i_{can}$ (\textit{``general task'', ``indistinct movement'', ``unclear action'', ``background''}). We manually annotate affordance segmentation masks for five affordable actions per sequence, resulting $\approx 700$ masks. We report mIoU.

Table \ref{tab:aff_segm} demonstrates the effectiveness of video-language features in capturing actions. Previous methods that rely on single-image CLIP features miss the dynamic action context in egocentric videos.
Consequently, the video-language feature field of DIV-FF excels in the affordance segmentation, achieving 20.7 mIoU (+69.7 $\%$), benefiting from training on video narrations, unlike CLIP models that use static image captions. 
We visualize these differences in Figure \ref{fig:aff_seg}, showing the relevancy scores for text queries detailing specific actions.
The image-language model performs well when actions are explicitly linked to objects, such as \textit{``cut the onion''} or \textit{``add ingredients to the mixture''}.
\begin{figure}[t]
    \centering
    \includegraphics[width=0.99\linewidth]{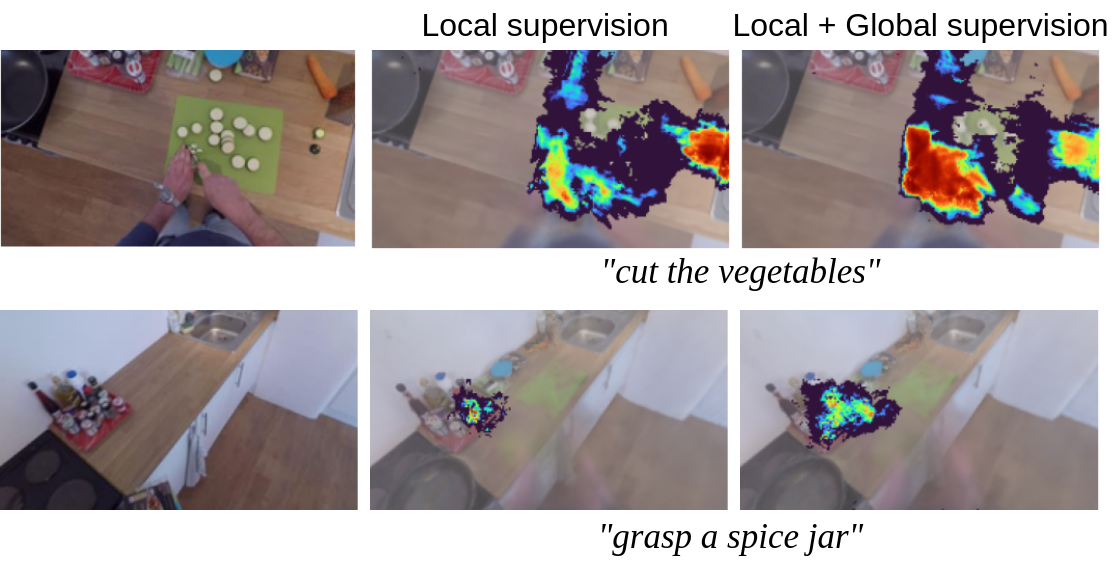}
    \caption{\textbf{Video-Language Loss ablation.} Including the global supervision term in the interaction hotspot mask produces sharper relevancy maps compared to just having the patch-level (local) term of the loss.}
    \label{fig:vid_abla}
\end{figure}
However, it struggles with verbs or semantic contexts that imply a location, like \textit{``wash a kitchen utensil''}—which suggests the sink—or \textit{``toast the bread''}, associated with the toaster. In these instances, the video-language model distinctly outperforms, accurately identifying the action's interaction hotspot. We also highlight that the localization of these fine-grained areas is due to the additional global supervision to the local medium-size Ego-Video patch features, as Figure \ref{fig:vid_abla} shows. The joined effect of the two losses improves the relevancy maps by explicitly guiding the optimization toward the interaction hotspot regions. Table \ref{tab:aff_segm} also shows that for single-image models, patch-based methods (LERF, CLIP in patches) outperform the full model using object masks, as we suggested in Section \ref{sec:vlff}.

\subsection{Amodal Scene Understanding.}

\begin{figure}[t]
    \centering
    \includegraphics[width=0.99\linewidth]{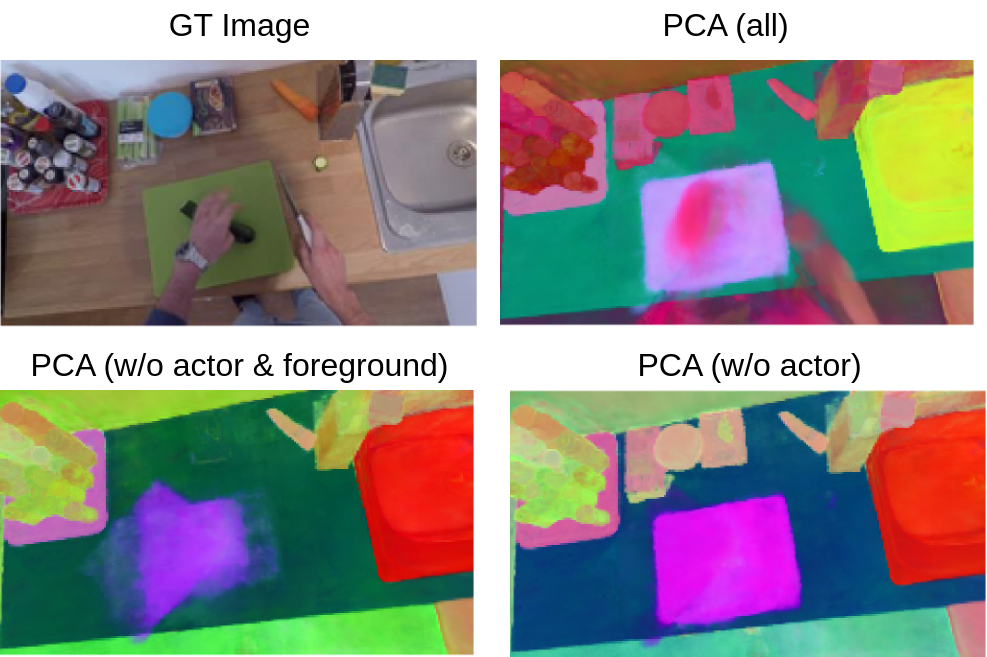}
    \caption{\textbf{Amodal Scene Understanding.} We visualize the PCA components obtained from the different composition of the image-text feature fields, showing accurate decomposition of the objects contours due to the SAM masks regularizing effect.}
    \label{fig:PCA}
\end{figure}

Our DIV-FF model comprises three distinct levels of geometry and semantics, representing different scene levels as illustrated in Figure \ref{fig:PCA}. 
This introduces significant versatility in the environment understanding.
For example, we can remove the actor's hands to reveal the dynamic objects without occlusions. 
Additionally, eliminating both the actor and dynamic elements exposes only the persistent parts of the scene. Our intuition is that this static spatial-semantic representation contains strong priors that can be exploited when the user revisits the scene at another time.

\section{Conclusions}

We proposed Dynamic Image-Video Feature Fields (DIV-FF) to address the limitations of existing egocentric video understanding methods. By decoupling the scene into persistent, dynamic, and actor streams and integrating video-based semantics, our approach achieves robust and consistent semantic segmentation over time. The model's ability to perceive and reason about both persistent and evolving scene elements marks a significant improvement in affordance and dynamic object understanding. Experimental results highlight DIV-FF's effectiveness in representing the rich and dynamic nature of egocentric environments, setting a promising direction for future work in spatial-temporal scene modeling and interaction-aware perception.

{
    \small
    \bibliographystyle{ieeenat_fullname}
    \bibliography{main}
}



\end{document}